\documentclass[sigconf]{acmart}
\usepackage[latin9]{inputenc}
\usepackage{textcomp}
\usepackage{graphicx}
\ifx\hypersetup\undefined
  \AtBeginDocument{%
    \hypersetup{unicode=true,
 bookmarks=false,
 breaklinks=false,pdfborder={0 0 1},backref=section,colorlinks=false}
  }
\else
  \hypersetup{unicode=true,
 bookmarks=false,
 breaklinks=false,pdfborder={0 0 1},backref=section,colorlinks=false}
\fi

\makeatletter

\newcommand{\noun}[1]{\textsc{#1}}
\DeclareTextSymbolDefault{\textquotedbl}{T1}

\AtBeginDocument{%
    \hypersetup{unicode=true}
  }%%
\AtBeginDocument{%
  }

\setcopyright{acmlicensed}
\copyrightyear{2025}
\acmYear{2025}
\usepackage{url}

\makeatother

\begin{document}
\title{Attribution analysis of legal language as used by LLM}
\author{Richard K. Belew}
\email{rbelew@ucsd.edu}
\begin{abstract}
Three publicly-available LLM specifically designed for legal tasks
have been implemented and shown that classification accuracy can benefit
from training over legal corpora, but why and how? Here we use two
publicly-available legal datasets, a simpler binary classification
task of ``overruling'' texts, and a more elaborate multiple choice
task identifying ``holding'' judicial decisions. We report on experiments
contrasting the legal LLM and a generic BERT model for comparison,
against both datasets. We use integrated gradient attribution techniques
to impute ``causes'' of variation in the models' perfomance, and
characterize them in terms of the tokenizations each use. We find
that while all models can correctly classify some test examples from
the casehold task, other examples can only be identified by only one,
model, and attribution can be used to highlight the reasons for this.
We find that differential behavior of the models' tokenizers accounts
for most of the difference and analyze these differences in terms
of the legal language they process. Frequency analysis of tokens generated
by dataset texts, combined with use of known ``stop word'' lists,
allow identification of tokens that are clear signifiers of legal
topics. 
\end{abstract}
\maketitle

\section{Introduction}

Large language models (LLM) have generally been constructed to consume
language of any sort. The specific language used in legal applications,
from judicial decisions to legislation to contracts, provides a key
opportunity to investigate how features of more constrained training
language samples on which LLM are applied can impact their performance.
The work presented here builds on two prior efforts to target LLM
on legal texts. The first are two models described by \citet{zhengguha2021},
one that begins with BERT, a well-known and language-generic model
that is then subsequently trained on legal texts, and one that is
trained exclusively on legal texts. A third legal model is provided
by \citet{chalkidis-etal-2020-legal} which was trained on a different
(but partially overlapping) legal corpus. Both groups have made the
exemplary effort of providing their trained models via the HuggingFace
public repository, supporting further investigations such as ours.
We focus on two classification tasks addressed in \citet{zhengguha2021},
overruling and holding. A first step in our work is to demostrate
that we have replicated performance of the \citet{zhengguha2021}
models on these tasks quite exactly.

Beyond just comparing accuracy of the models on shared tasks, the
important questions now have to do with \emph{explaining why} the
various models act as they do. ``Explainable AI'' (XAI) has become
an increasingly important topic to everyone concerned with making
the models' behavior sensible to human users, to generate trust in
their results, as well as providing insights for more sophisticated
models \citep{bennetot2022practicalguideexplainableai}. There have
been a variety of ``attribution'' methods proposed to identify,
within the many network modules and layers and millions/billions of
network parameters, just what is it ``causing'' the model to produce
the outputs it does in response to the inputs it has been given \citep{madsen22}.
We focus here on ``integrated gradient'' attribution methods \citep{sundararajan2017axiomatic}.
Our central question is: what can we learn using attribution methods
over the legal LLM to understand why each of them behave as they do,
and how their behaviors differ one from another.

\section{Methods}

\subsection{LLM models}
\begin{enumerate}
\item \noun{BERT} (double) model\footnote{https://huggingface.co/google-bert/bert-base-uncased}:
\citet{zhengguha2021} began with standard BERT model \citet{devlin2019bert}
base model, which had been trained over English Wikipedia corpus for
1M steps using the WordPiece \citep{schuster2012japanese} tokenizer.
It was then trained for an additional 1M steps over the same Wikipedia
corpus, using the MLM {[}masked language model{]} and NSP {[}next
sentence prediction{]} objectives. 
\item \noun{legalBERT}\footnote{https://huggingface.co/casehold/legalbert}:
\citet{zhengguha2021} begin with the standard BERT base model, but
it is then trained: 
\begin{quotation}
... over the Harvard Law \noun{case.law} \citep{Caselaw_harvard}
corpus from 1965 to the present. The size of this corpus (37GB) is
substantial, representing 3,446,187 legal decisions across all federal
and state courts... for an additional 1M steps on the MLM and NSP
objective, with tokenization and sentence segmentation adapted for
legal text. 
\end{quotation}
A subset of the decisions was held out for use for fine-tuning and
testing for the casehold task. 
\item \noun{customLegalBERT}\footnote{https://huggingface.co/casehold/custom-legalbert}:
This model is ``pretrained from scratch for 2M steps using {[}just{]}
the case law corpus and has a custom legal domain-specific vocabulary'':
``The vocabulary set is constructed using SentencePiece \citep{kudo2018sentencepiece}
on a subsample (appx. 13M) of sentences from our pretraining corpus,
with the number of tokens fixed to 32,000.'' 
\item \noun{nlpaueb}\footnote{https://huggingface.co/nlpaueb/legal-bert-base-uncased}:
(It so-named here because of its unfortunately similar name ``Legal-BERT''
to the model by \citet{zhengguha2021}): This model is built from
``12 GB of diverse English legal text from several fields (e.g.,
legislation, court cases, contracts) scraped from publicly available
resources. `` Notably, its training corpus also include 164141 cases
taken from the \noun{case.law} corpus. nlpaueb also uses the SentencePiece
\citep{kudo2018sentencepiece} tokenizer, constrained to the same
vocabulary of size 30522. 
\end{enumerate}
The models have been ordered in terms of the amount of legal text
included in their training corpora, from none in \noun{BERT} to exclusively
legal in \noun{customLegalBERT} and \noun{nlpaueb}, because this will
be shown to be an important dimension to their varying performance.
All models use the same number of parameters (110M).

\subsection{Datasets}

\subsubsection{Overrule}

\begin{figure}
\begin{centering}
\includegraphics[width=0.9\columnwidth]{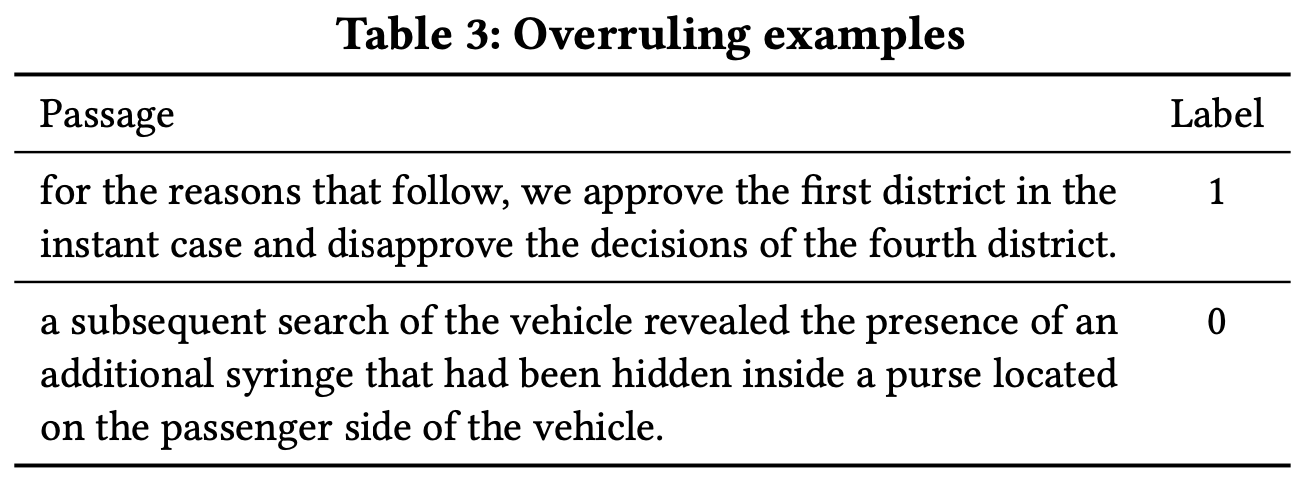} 
\par\end{centering}
\caption{\label{fig:overruleEG}Positive and negative overruling examples (Table
3 from \citeauthor{zhengguha2021})}
\end{figure}

Both the \noun{overrule} and \noun{casehold} datasets are best described
by \citet{zhengguha2021}. In brief, \noun{overule} is a binary classification
task where sentences from judicial opinions either do or do not rule
so as to change prior law. The dataset was produced by Casetext.com,
with positive examples identified by \emph{manual} annotation, by
attorneys, and randomly selected negative examples. Positive and negative
examples provided by \citet{zhengguha2021} are reproduced as Figure
\ref{fig:overruleEG}.

Like many legal topics, the action of judicial overruling is rich
and so the language judges use will be, too. \citet{dunn2003judges}
has described a subtle process in the judges' writing: ``They occupy
a paradoxical position in this world, one in which their function
requires them to make law, while their legitimacy depends on the fiction
that they interpret law.''

Dunn focuses on specific lexical elements and relates them to speech
act analysis. Specific words and phrases from his analysis will be
referenced later; cf. Section \ref{subsec:Connecting-LLM-to-legal-language}.
Despite the linguistic subtlety he describe, it turns out (cf. Section
\ref{subsec:Classification-performance}) LLMs can do very well at
making this binary distinction.Fine-tuning training of all models
was done over 2000 examples, then tested over a separate sample of
400 examples. 

\subsubsection{Casehold\label{subsec:CaseholdDS}}

\begin{figure}
\begin{centering}
\includegraphics[height=4in]{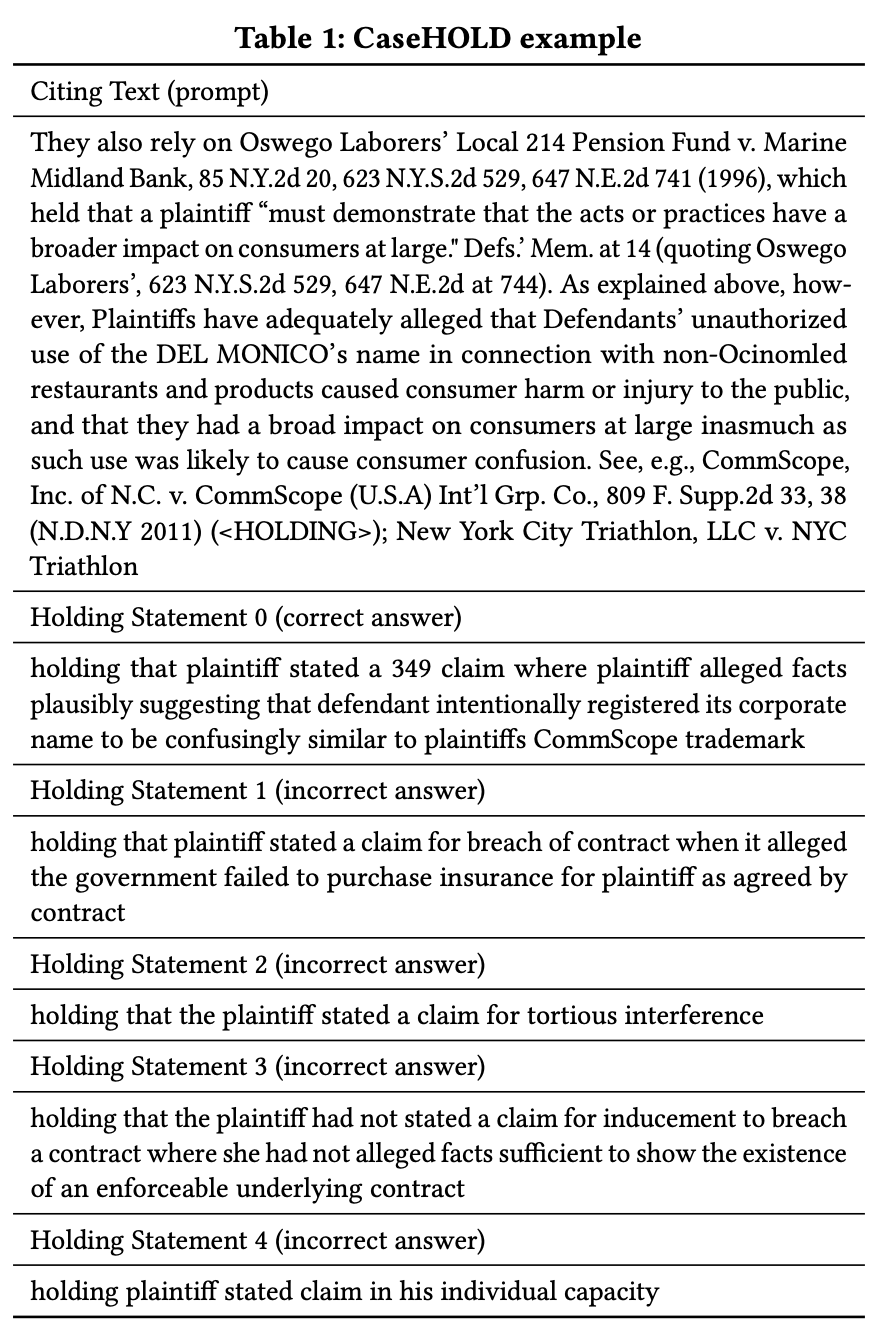}
\par\end{centering}
\caption{\label{fig:casehold-example}casehold example (Table 1 from \citet{zhengguha2021})}
\end{figure}

The \noun{casehold} dataset is a multiple choice task of the sort
first defined by the SWAG experiments \citep{zellers2018swag}. \citet{zhengguha2021}
describe the \noun{casehold} version: 
\begin{quotation}
The citing context from the judicial decision serves as the prompt
for the question. The answer choices are holding statements derived
from citations following text in a legal decision. There are five
answer choices for each citing text. The correct answer is the holding
statement that corresponds to the citing text. The four incorrect
answers are other holding statements. 
\end{quotation}
An example presented by \citet{zhengguha2021} is shown in Figure
\ref{fig:casehold-example}.

Care was taken to have the four incorrect answers be plausible, by
computing text similarity between the correct answer and alternative
statements extracted from the corpus. Fine-tuning training of all
models was done over 42509 examples, then tested over a separate sample
of 5314 examples.

\subsection{Attribution across LLM components}

\begin{figure}
\begin{centering}
\includegraphics[width=0.8\columnwidth]{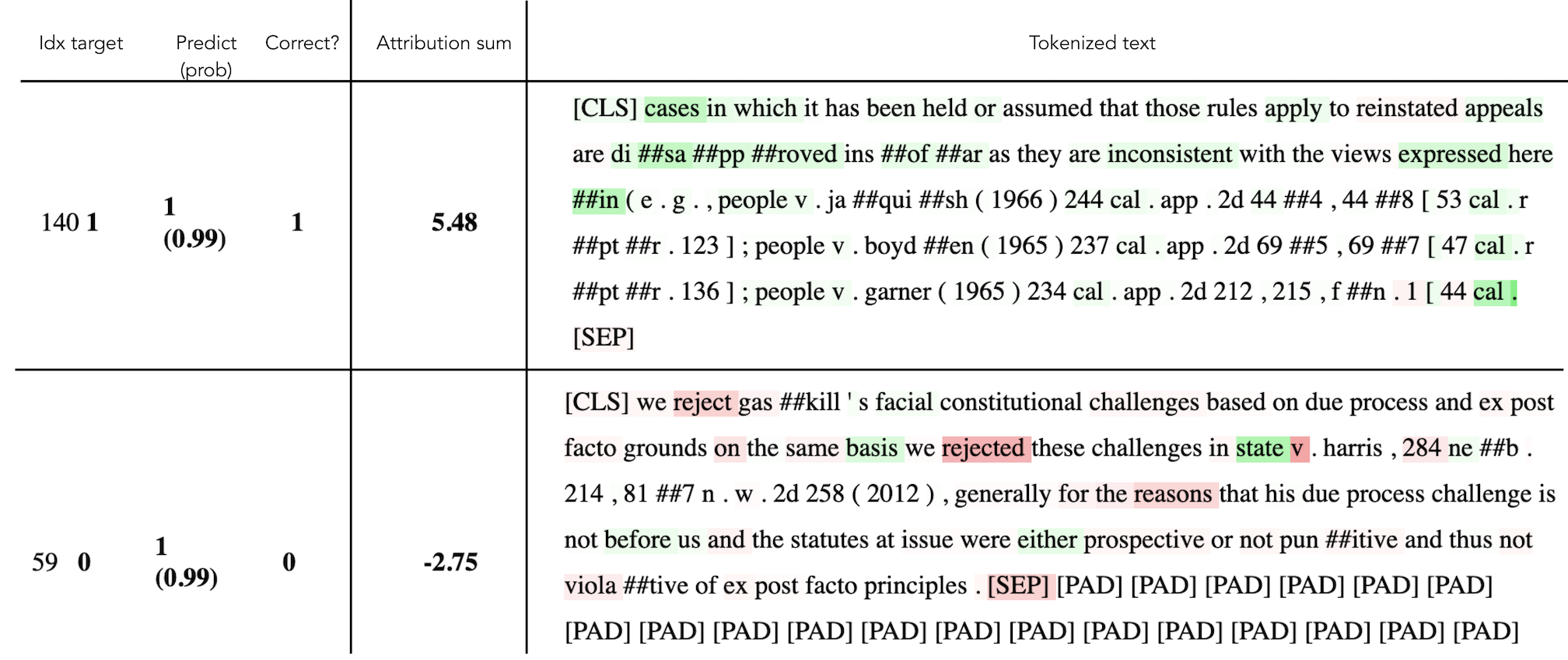} 
\par\end{centering}
\caption{\label{fig:attrib_samples}Sample attributions (\noun{overrule, legalBERT})}
\end{figure}

As deep neural network models continue to become more elaborate and
the inferences they produce continue to expand, the desire to ``apportion
credit'' across network components via ``explainable AI'' (XAI)
techniques continues to grow. Early ``gradient'' methods \citep{simonyan2013deep}
exploited the same partial derivatives of inputs with respect to target
values driving the back propagation learning algorithms that establish
network weights. Intuitively, finding large gradients in weight change
with respect to a (word/token) unit reflects the ``salience'' of
its weight changes in model error. Integrated gradients \citep{sundararajan2017axiomatic}
extends this approach with path integrals: 
\begin{quotation}
Gradients (of the output with respect to the input) is a natural analog
of the model coefficients for a deep network, and therefore the product
of the gradient and feature values is a reasonable starting point
for an attribution method.... Specifically, let x be the input at
hand, and x\textasciiacute{} be {[}a{]} baseline input. ... Integrated
gradients are defined as the path intergral of the gradients along
the straightline path from the baseline x\textasciiacute{} to the
input x. 
\end{quotation}
This technique as implemented in Captum\footnote{https://captum.ai}:
\begin{quotation}
...allow us to assign an attribution score to each word/token embedding
tensor in the {[}movie review{]} text. We will ultimately sum the
attribution scores across all embedding dimensions for each word/token
in order to attain a word/token level attribution score. \citep{captumTutorailIMDB} 
\end{quotation}
Figure \ref{fig:attrib_samples} shows positive (green) and negative
(red) attributions across tokens for two examples from the overrule
dataset as classified by the legalBert model: a correctly labeled
example with primarily positive attributions above, and an incorrectly
labeled example with primarily negative attributions below. Also listed
are the prediction probabilities and attribution sums for each example
across all tokens.

Below we describe the results of collecting per-token statistics across
all dataset samples.

\section{Results}

\subsection{Classification performance\label{subsec:Classification-performance}}

\begin{figure}
\begin{centering}
\includegraphics[height=1.5in]{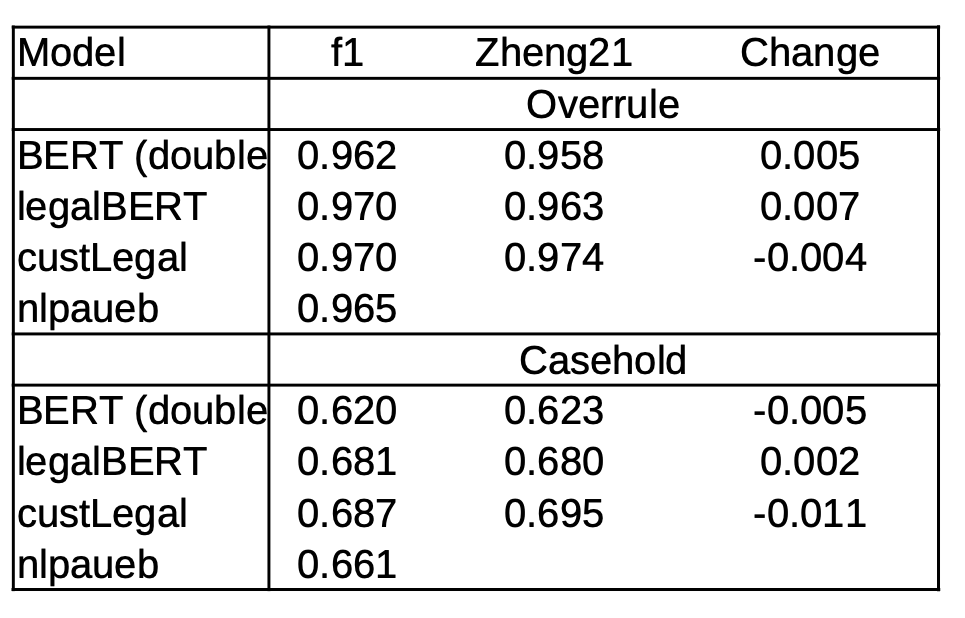} 
\par\end{centering}
\caption{\label{fig:Model-performance}Model performance}
\end{figure}

The basic classification results are shown in Figure \ref{fig:Model-performance}.
With respect to the \noun{overrule} data set, all models produced
classification accuracy scores above 95\% as measured by the f1 statistic.
The performance of models re-implemented for the experiments reported
here are also compared to those reported by \citet{zhengguha2021}
and demonstrate accurate replication of those prior results.

With respect to the \noun{casehold} data set, all models' performance
was lower than on the simpler overrule dataset, but broadly consistent
across models. Plain \noun{BERT} did least well and the two models
of \citet{zhengguha2021} did better than those than the \noun{nlpaueb}
model. Again, the models implemented here produce results very similar
to those given by \citet{zhengguha2021} 

\subsection{Variations in attribution\label{subsec:Variations-in-attribution}}

\begin{figure}
\begin{centering}
\includegraphics[height=1.5in]{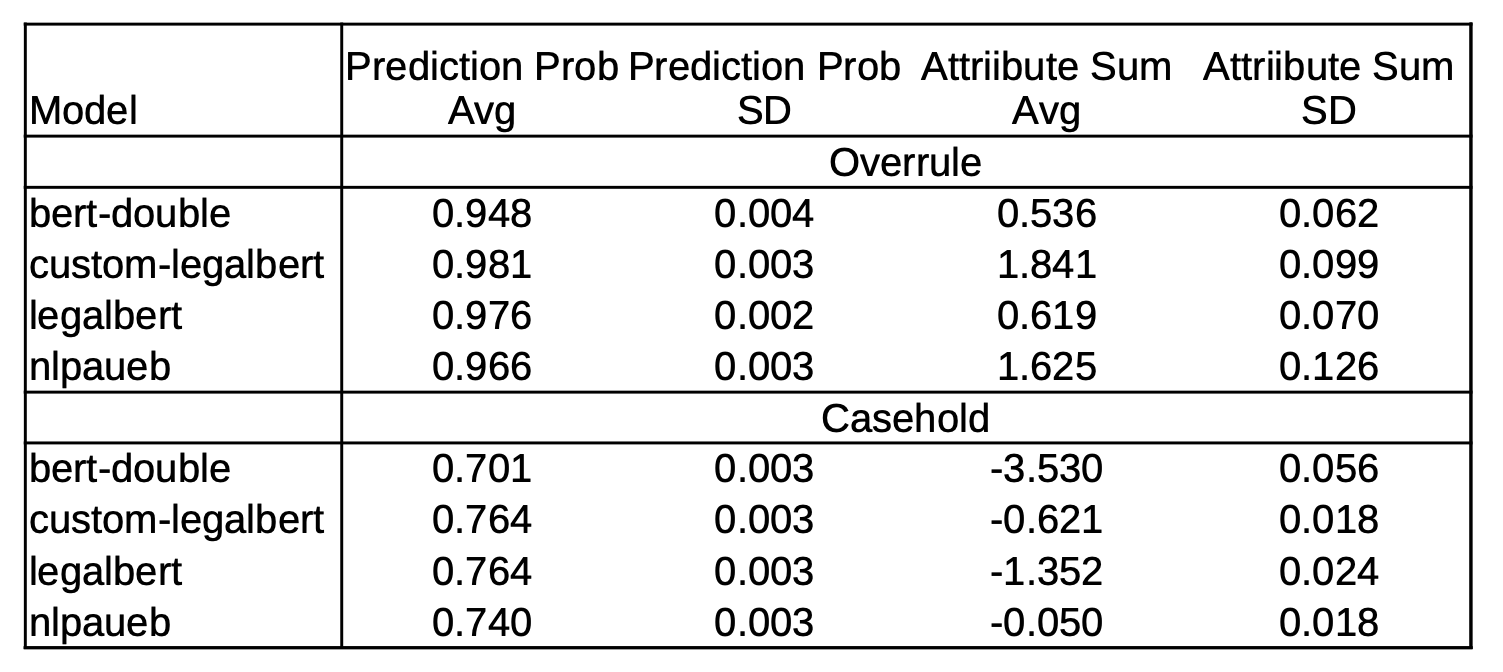} 
\par\end{centering}
\caption{\label{fig:distribStats}Prediction, attribution distribution}
\end{figure}

The statistics shown in Figure \ref{fig:distribStats} relate the
models' prediction probability averages and standard deviations across
all (testing) examples, and also the average and standard deviations
of attribution sums across example. For both \noun{overrule} and \noun{casehold}
data sets, prediction probability across models showed the same general
pattern, for example showing that on the \noun{casehold} dataset BERT
average prediction probability was lower than \noun{nlpaueb}, and
that both were lower than the two \citet{zhengguha2021} models.

The level of attribution applied to examples varied much more across
both datasets and models. On the \noun{overrule} dataset, the \noun{customLegalBert}
and \noun{nlpaueb} models produced much greater \emph{positive} attribution
than \noun{BERT} or \noun{legalBert}. Yet on the casehold dataset
the situation was exactly reversed: very \emph{negative} average attributions
were generated by \noun{BERT} and \noun{legalBert}.

One interesting trend in attribution scores is that they roughly track
the degree to which the models depended on training over legal corpora:
\noun{BERT} had the least, \noun{legalbert} had a mix of \noun{BERT's}
training with subsequent legal text training, and \noun{customLegalBert}
and \noun{nlpaueb} were trained on exclusively legal texts.

\begin{figure}
\begin{centering}
\includegraphics[width=0.8\columnwidth]{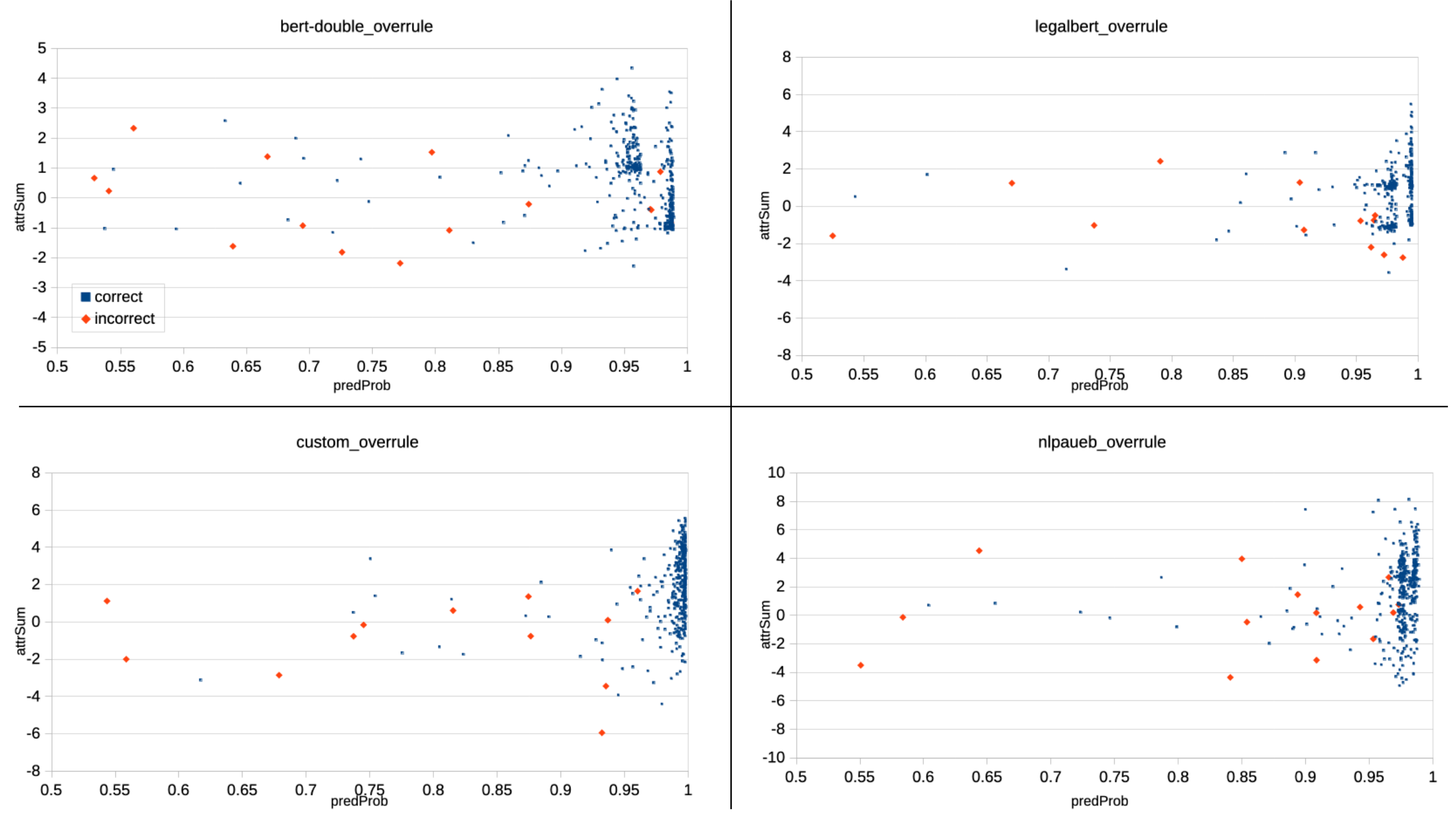} 
\par\end{centering}
\caption{\label{fig:predVattr_overrule}Prediction vs attribution - Overrule}
\end{figure}

Some insight can be obtained by scattering prediction probabilities
versus attribute sums. Figure \ref{fig:predVattr_overrule} shows
this for the four models on the overrule dataset. As evidenced by
the high classification scores, all models produced vertical stripes
near the prediction=1.0 value, for all models except \noun{customLegalBert}.
The two stripes seem to indicate nearly discrete prediction values
for two sets of examples. Note especially in the \noun{BERT} model's
graph that the attribution sums associated with one stripe begin at
a value of -1.0, whereas the second stripe begins with a +1.0 attribution
score. What should cause this discretization in probability scores,
and their connection to the valence of attribution score, is not clear. 

\begin{figure}
\begin{centering}
\includegraphics[width=0.8\columnwidth]{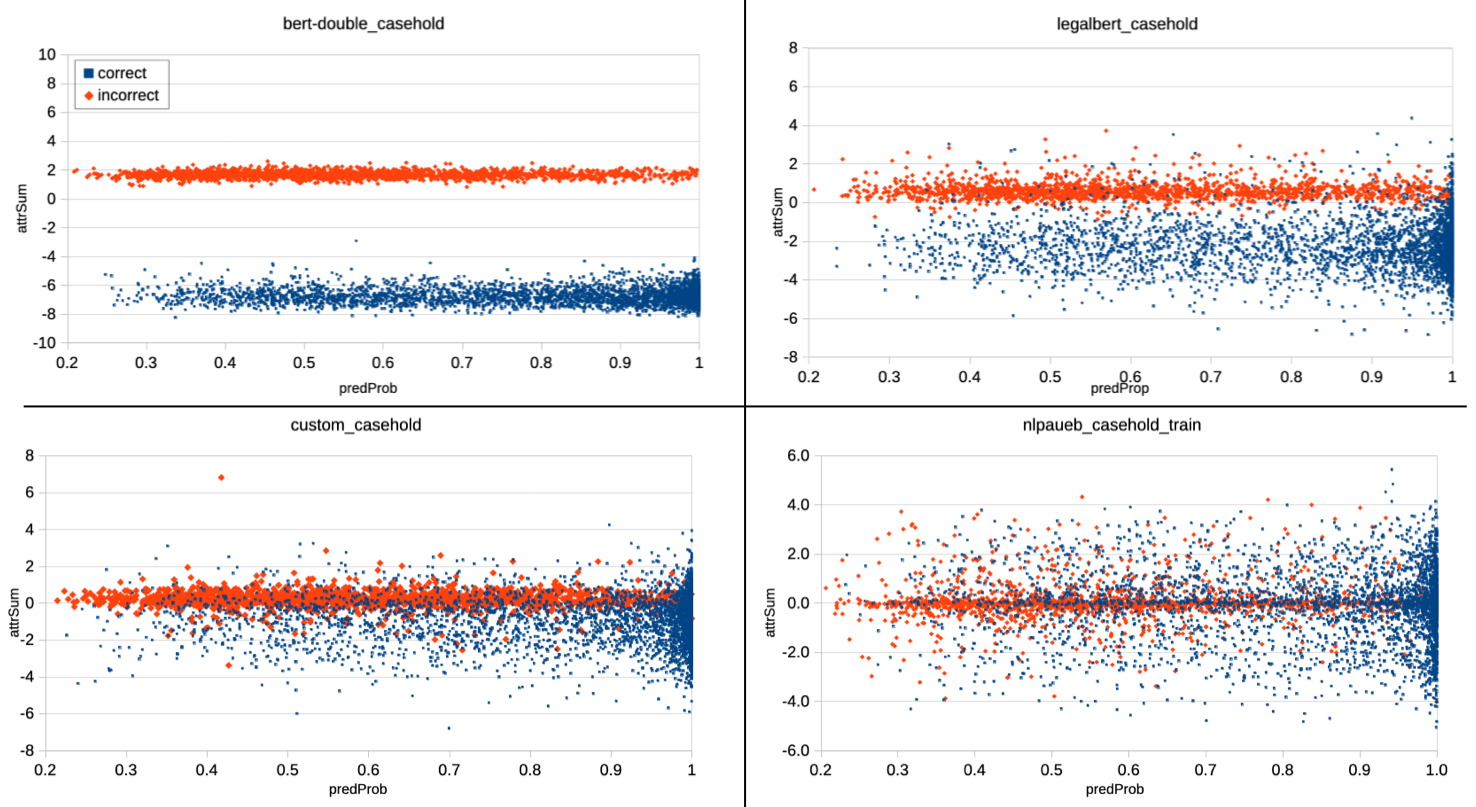} 
\par\end{centering}
\caption{\label{fig:predVattr_casehold}Prediction vs attribution - Casehold}
\end{figure}

Figure \ref{fig:predVattr_casehold} shows the same distributions
with respect to the casehold data set. Again, there is heavy weighting
of correct examples near the probability prediction=1.0. But the distribution
of correct versus incorrect examples with respect to attribution scores
is striking. Both \noun{BERT} and \noun{legalBert} show clear distinctions
between average attribution scores of correct versus incorrect examples.
As mentioned above, the trend in attribution scores roughly track
the degree to which the models depended on training over legal corpora. 

Classifier labels are provided as input to calculation of attribution
scores, and the scores' valence is a consequence of the gradient's
direction being reversed for correct and incorrect examples. Models
generally produce negative attribution for correctly classified instances
and positive attribution for incorrect ones. This is very true using
the \noun{BERT} model, quite true of \noun{legalBert}, and barely
true of \noun{customLegalBert}.Attribution scores using the \noun{nlpaueb}
model are quite uniformly distributed

\subsection{Meanings in the vocabularies}

\subsubsection{Variation across LLM vocabularies\label{subsec:Variation-across-LLM}}

Independent of classification performance, we can consider how the
various LLM perceive the inputs on which they are trained. In fact,
because the models all share the same WordPiece techniques, variations
in their vocabularies is a direct result of the corpora over which
they were trained.

\begin{figure}
\begin{centering}
\includegraphics[width=0.8\columnwidth]{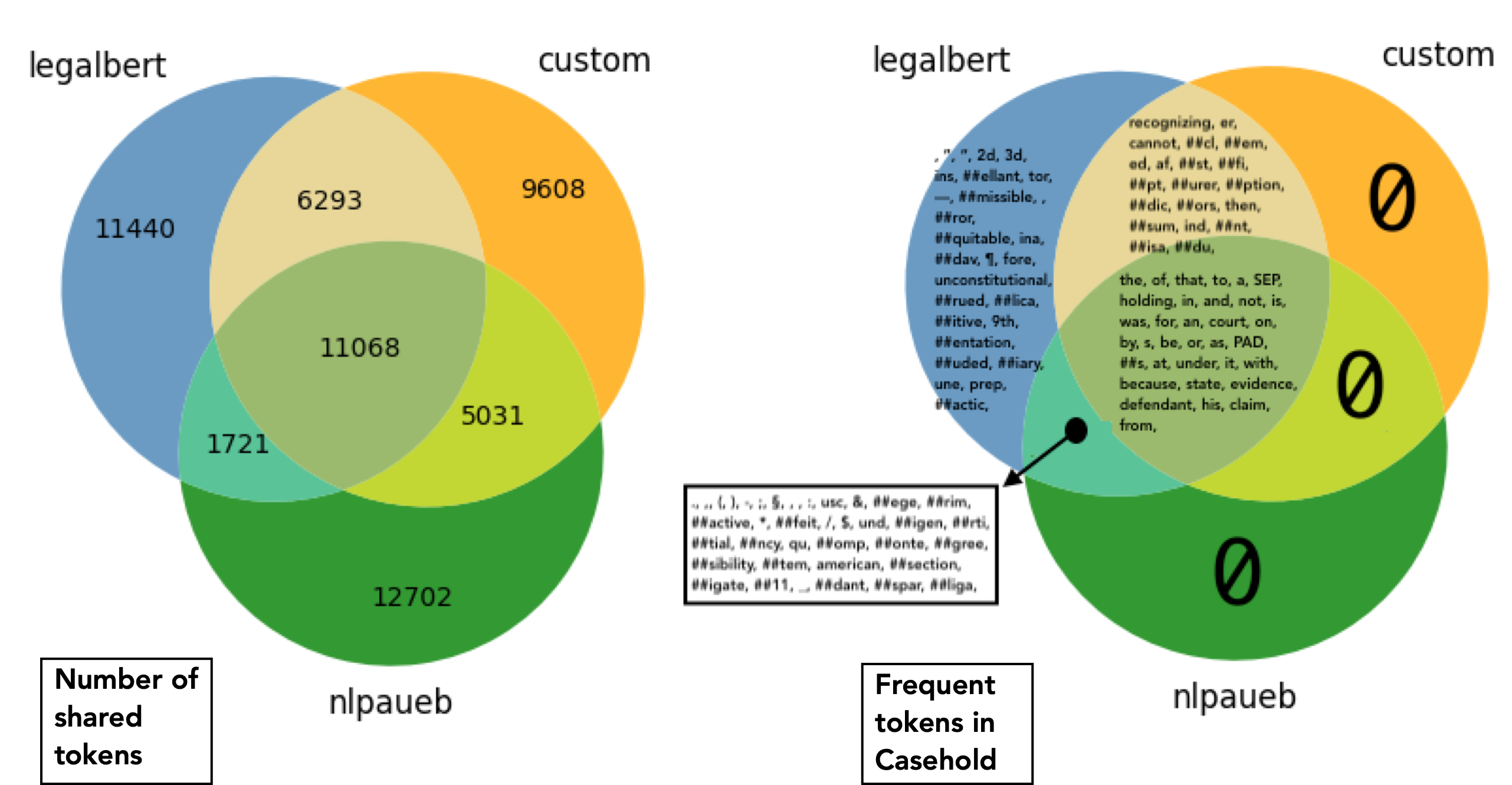} 
\par\end{centering}
\caption{\label{fig:vocabOverlap}Vocabulary overlaps}
\end{figure}

Figure \ref{fig:vocabOverlap} shows first the size of the token subsets
shared across the models. There is considerable variation: the union
across the vocabularies has 57863 unique tokens. Shown next is an
analysis of the overlaps in terms of the ``words'' (decoded tokens)
filtered to show the \emph{50 most frequent} words occuring across
the entire\footnote{Because it plays no role in the models' training, analysis of frequent
terms is across the entire casehold dataset, not just training or
test splits. } casehold training dataset.

\begin{figure}
\begin{centering}
\includegraphics[height=4in]{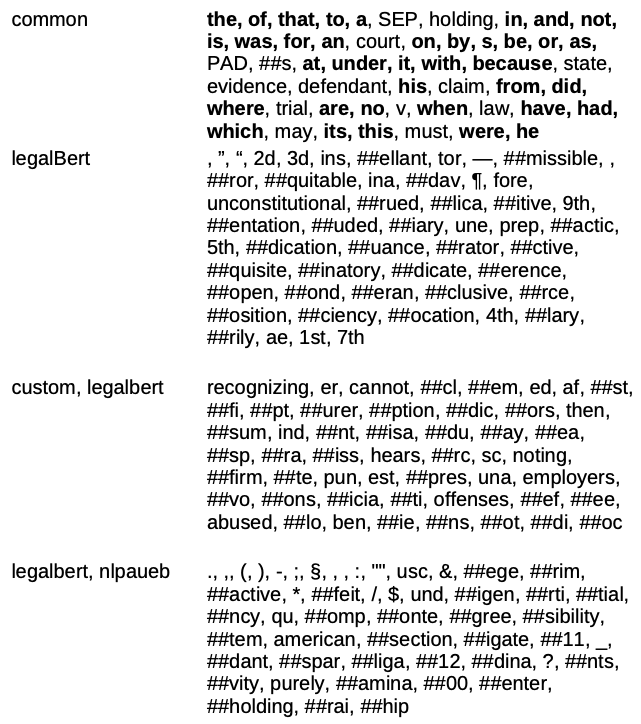} 
\par\end{centering}
\caption{\label{fig:vocabSet-enum}Tokens identified in various models' tokens
subsets}
\end{figure}

The words associated with each subset have been listed more legibly
in Figure \ref{fig:vocabSet-enum}. It is not surprising that many
of token/words shared by all models correspond closely to ``stop
words'' often identified by information retrieval techniques. Those
included in the NLTK \citep{bird2009natural} stop word list have
been listed in boldface.The common tokens \emph{not} part of the stop
word list are also interesting, as clear signifiers of legal topics.

\begin{figure}
\begin{centering}
\includegraphics[height=4in]{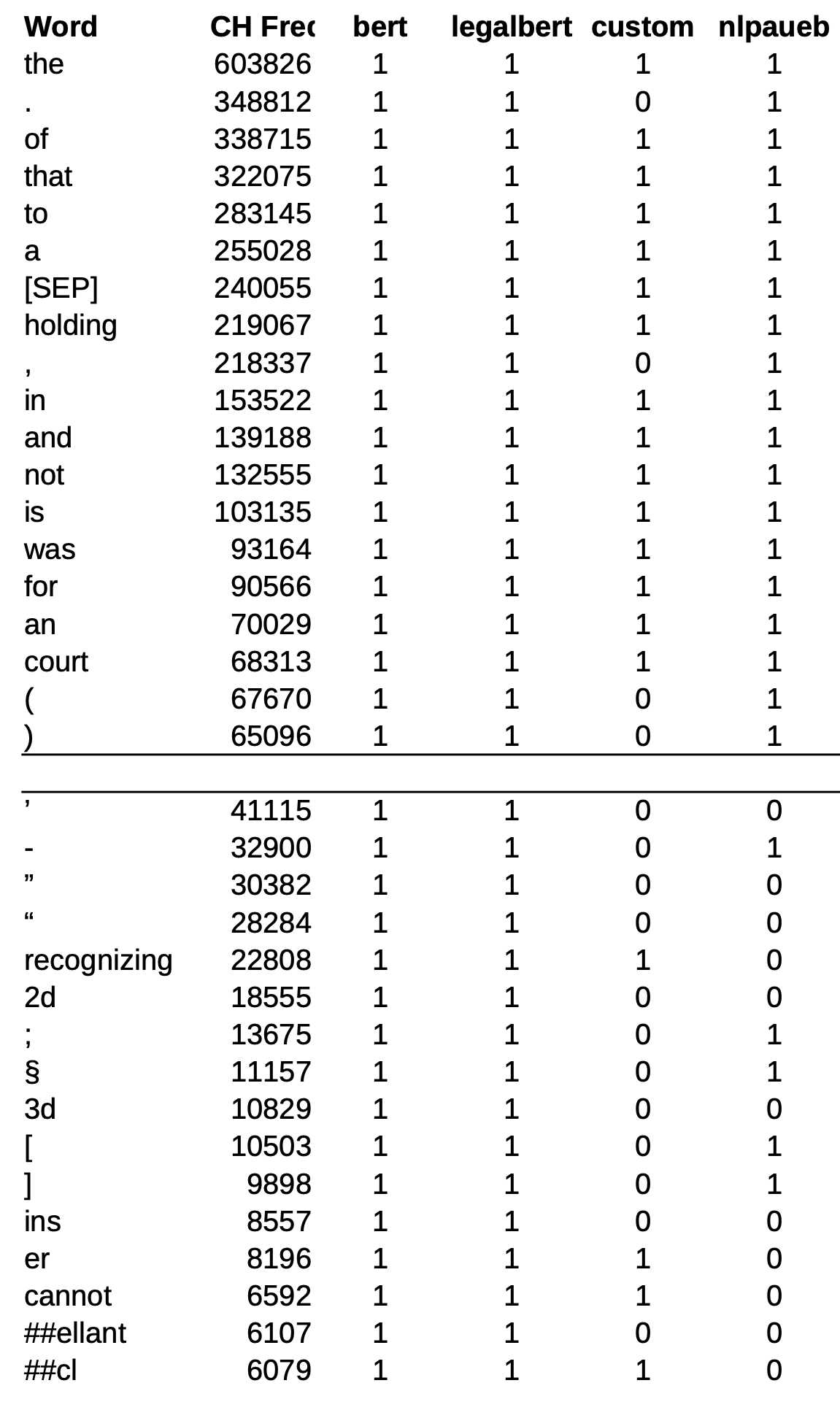} 
\par\end{centering}
\caption{\label{fig:modelData_overl;ap}Models' tokens' overlap on casehold }
\end{figure}

Figure \ref{fig:modelData_overl;ap} shows frequent tokens from the
\noun{casehold} dataset together all with the models including them.
All of the 20 most frequent tokens are included, then a sample of
other tokens captured/missed by various models' tokenizers. Missed
tokens are often, but not exclusively, from punctuation: \noun{recognizing}
is missing from \noun{nlpaueb}, \noun{ins} and \noun{\#\#ellant }from
both \noun{nlpaueb} and \noun{customLegalBert}. Every token within
the \noun{casehold} dataset is captured by the \noun{BERT} model and
also the \noun{legalBert} model, because they share the same tokenizer.

\begin{figure*}
\begin{centering}
\includegraphics[width=0.8\paperwidth]{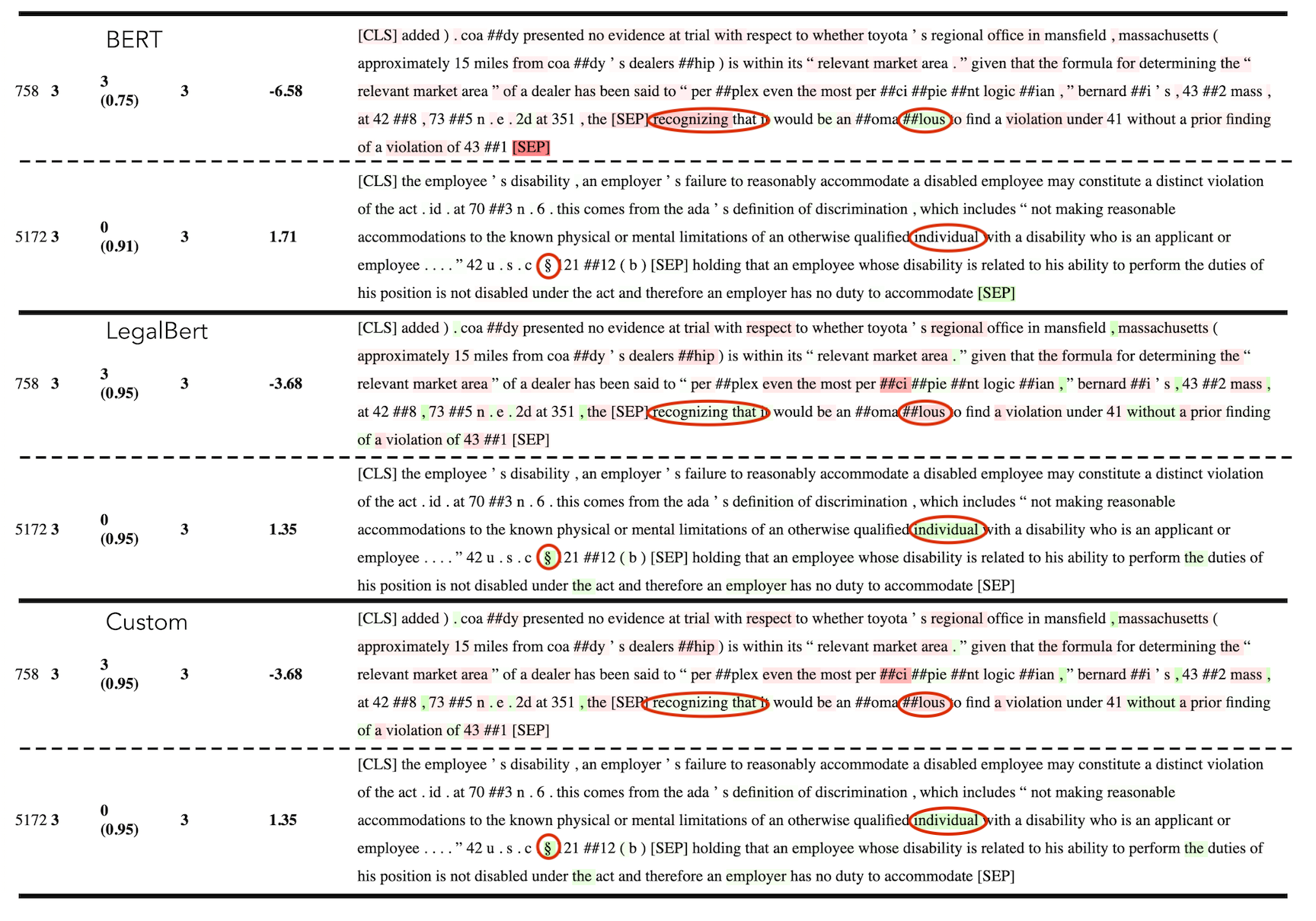}
\par\end{centering}
\caption{\label{fig:varyingClassif}Varying model attributions of two samples
by legal models}
\end{figure*}

These differences have consequence with respect to the models' classification.
Figure \ref{fig:varyingClassif} shows a pair of casehold examples
as classified by \noun{BERT, legalBert} and \noun{customLegalbert}
models, one correctly (\#0758) and another misclassified (\#5172).
The same attribution color coding of tokens is applied as in Figure
\ref{fig:attrib_samples}, and several tokens with varying attributions
have been highlighted. Some of these are among the list of most frequent
tokens just listed.

\begin{figure}
\begin{centering}
\includegraphics[height=2in]{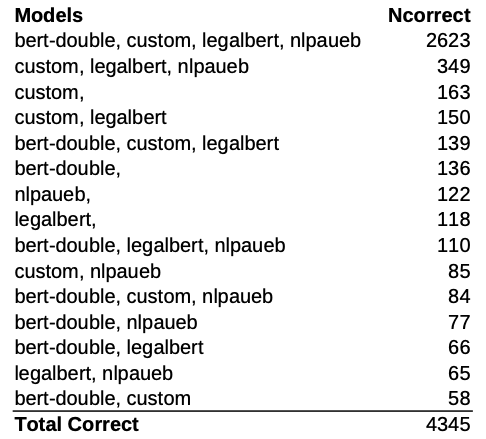}
\par\end{centering}
\caption{\label{fig:correctEG_sets}Number of examples correctly classified
by various models}
\end{figure}

Figure \ref{fig:correctEG_sets} summarizes the set of examples correctly
classified by one or more of the models. More than half of the correctly
classified examples are so classified by all four models, but there
are other examples correctly classified only by various sets of the
models. Given that fully 4345 of the 5314 testing examples were correctly
classified by one or more models, it is conceivable that ``mixture
of expert'' techniques could combine them to attain accuracies near
80\%.

\begin{figure*}
\begin{centering}
\includegraphics[width=0.8\paperwidth]{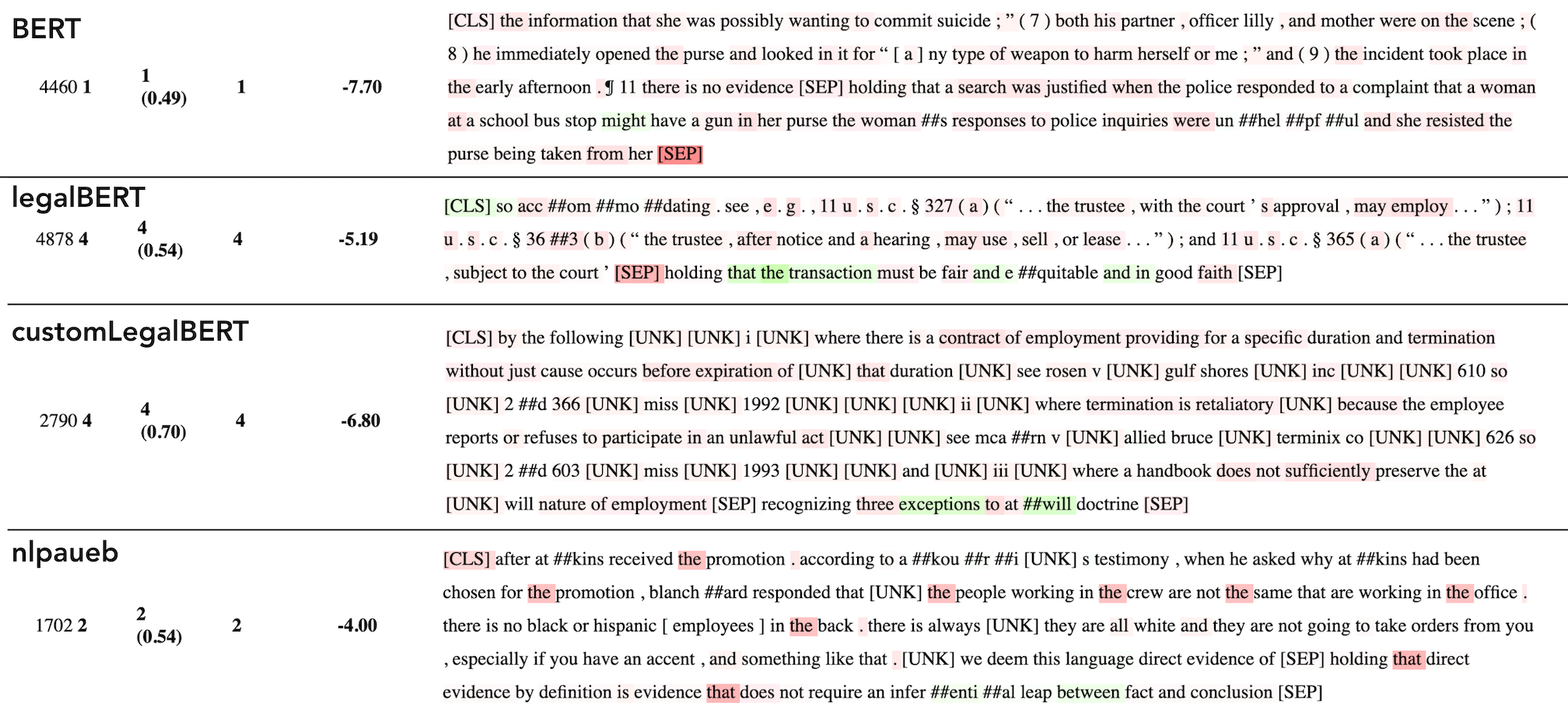}
\par\end{centering}
\caption{\label{fig:correctEG}Attributed examples classified correctly by
only one of the models}
\end{figure*}

Figure \ref{fig:correctEG} shows a second set of examples selected
because they were correctly classified by only one of the four models.
All these examples were classified with only weak prediction probabilities.
All but \noun{BERT} have a mix of some postive attributions with predominantly
negative attribution probabilities.

\subsubsection{Connecting LLM with known legal language\label{subsec:Connecting-LLM-to-legal-language}}

While the tokenization methods (common to all models) do not make
linguistic analysis of their tokenized products straight-forward,
the analysis above shows that some semantic interpretation is possible.

\begin{figure}
\begin{centering}
\includegraphics[height=1.5in]{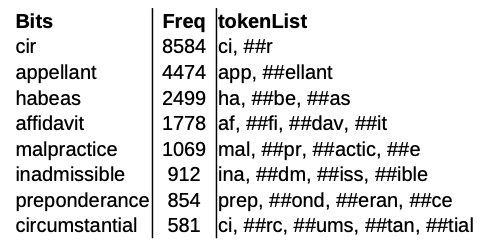} 
\par\end{centering}
\caption{\label{fig:brokenWords} Words formed from merged tokens}
\end{figure}

One simple trick unites ``broken words'': words in the text broken
by WordPiece (i.e., left with \noun{``\#\#''} prefixes on all-but-the-first
tokens). Figure \ref{fig:brokenWords} shows a series of the resulting
merged words, sorted by frequency in the \noun{casehold} dataset,
and comprising as many as five tokens.

\begin{figure}
\begin{centering}
\includegraphics[height=2in]{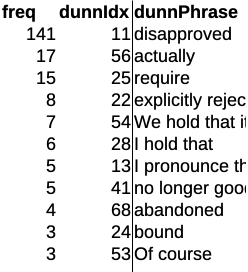} 
\par\end{centering}
\caption{\label{fig:dunn_overrule}Dunn phrases in overule}
\end{figure}

It is also possible to use prior legal scholarship as probes into
the LLM vocabulary. The work by \citet{dunn2003judges} highlighting
specific lexical elements and their relation to speech act theory
as used in overruling judicial decisions was mentioned above. A manually
curated list of 71 specific phrases mentioned in his article was assembled
and tokenized using the tokenizer (specifically, the WordPiece one
shared by \noun{BERT, legalBert}) to produce a set of token sequences.
Each of these can then be used as \emph{queries} against various datasets:
do the dataset example texts contain Dunn's phrases? Since Dunn's
interest was in the act of judicial overrule, we focus on results
against the \noun{overrule} dataset. Figure \ref{fig:dunn_overrule}
shows the most frequent mentions of Dunn's multi-token words and phrases
across the overrule dataset.

\subsubsection{Token frequencies across examples}

\begin{figure*}
\begin{centering}
\includegraphics[width=0.8\paperwidth]{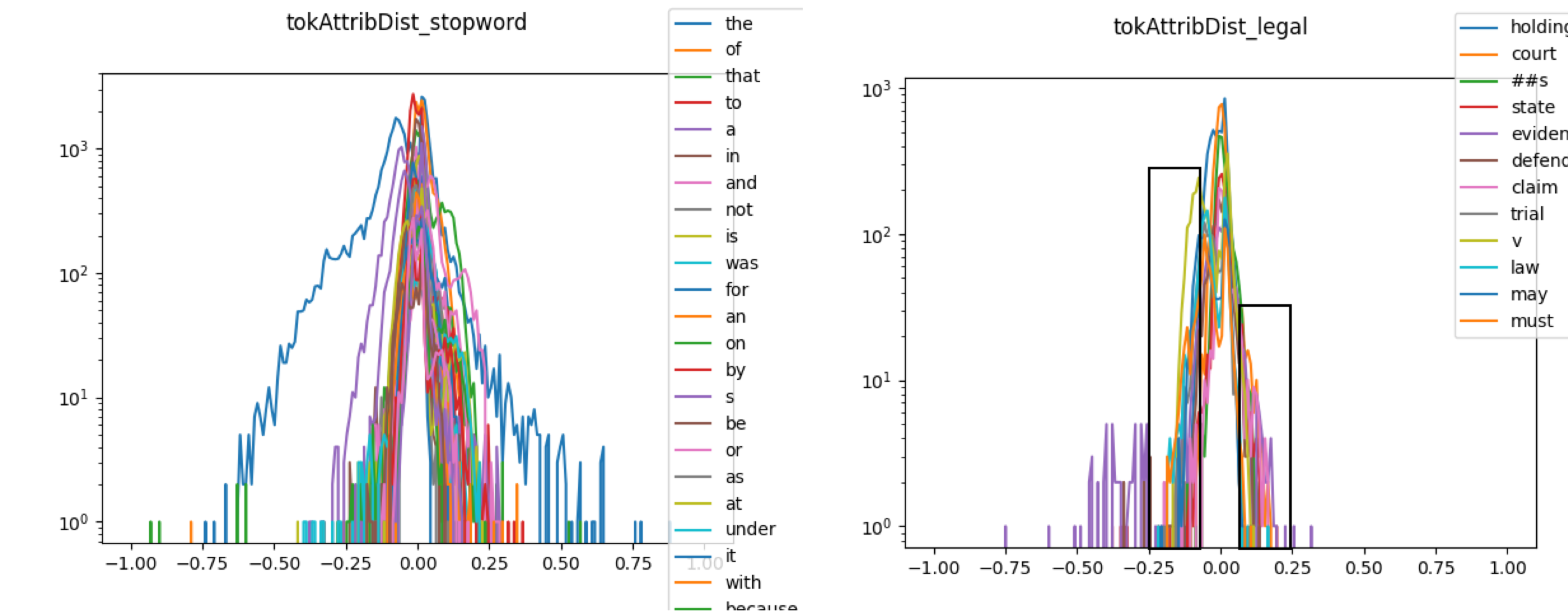}
\par\end{centering}
\caption{\label{fig:tokAttribDist}Token frequencies, stopwords and legal}
\end{figure*}

It seems reasonable to ask how the attribution assigned to tokens
in individual examples varies across the examples. Figure \ref{fig:tokAttribDist}
shows these distributions for the two sets of frequent tokens highlighted
in Figure \ref{fig:vocabSet-enum}, stopwords and legal (i.e., non-stopword)
tokens. Attribution scores have been binned at a resolution of 0.01.,
and the token frequency (vertical) axis is logarithmic. By far the
majority of all token attribution scores are near zero. There is a
clear skew towards negative attributions, for both stopwords and legal
tokens, especially evident with very common tokens ``\noun{the}''
and ``\noun{a}''.

\begin{figure*}
\begin{centering}
\includegraphics[width=0.8\paperwidth]{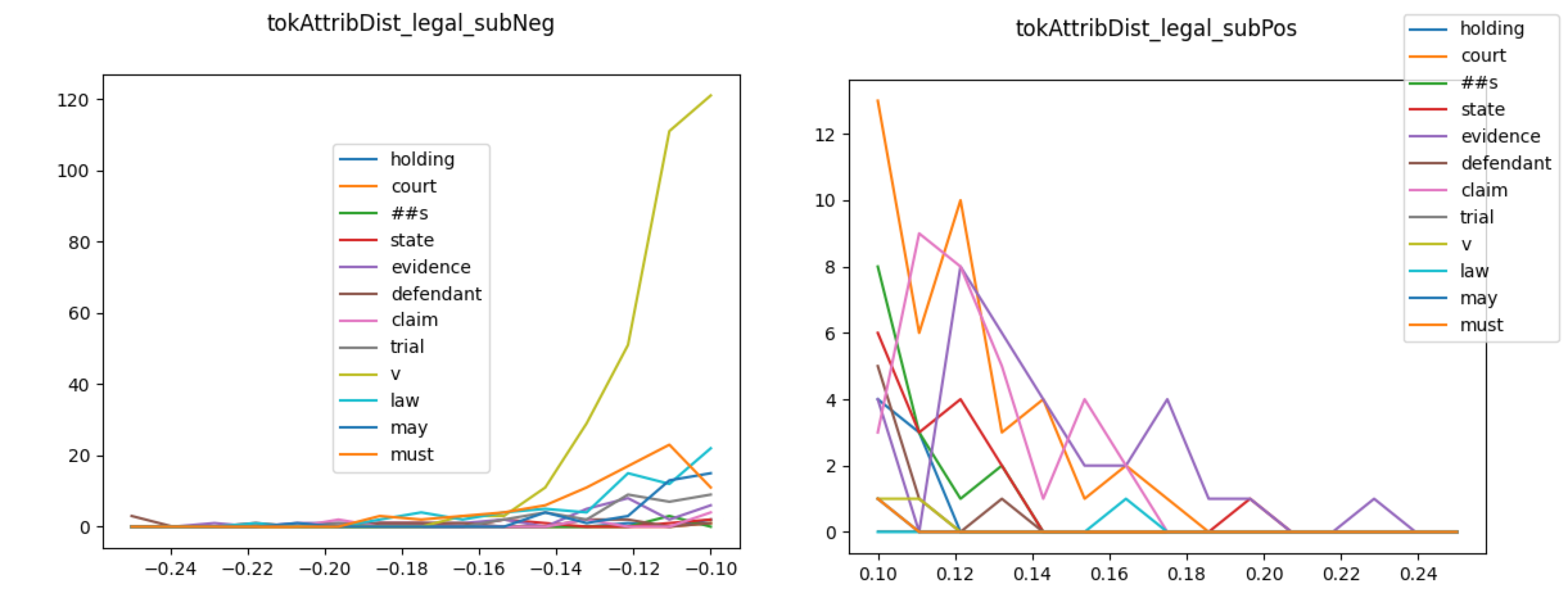}
\par\end{centering}
\caption{\label{fig:legalTokDist}Legal token frequencies, negative and positive}
\end{figure*}

A bit more insight can be provided by focusing on restricted positive
and negative ranges {[}0.10,0.25{]}; these have been highlighted in
the plot of legal tokens and shown in higher resolution in Figure
\ref{fig:legalTokDist} These show the token ``\noun{v}'' often
has significant negative attribution, while ``\noun{court}'' and
other legal tokens are more balanced.

\section{Discussion}

\subsection{Related and future work}

The integrated gradient attribution methods used here were described
by \citet{sundararajan2017axiomatic} and made available as part of
the Captum \footnote{https://captum.ai/} model interpretation library.

The worked described in \citet{zhengguha2021} was a central resource
for our work.Not only did it introduce both the \noun{overrule} and
\noun{casehold} datasets and make three of the models described publicly
available but also demonstration code\footnote{https://github.com/reglab/casehold}.
\citet{chalkidis-etal-2020-legal} similarly made their \noun{nlpaueb}
model available for use here.

The Chalkidis group has continued pursue related work. In \citet{Chalkidis2023LeXFilesAL}
they compares not only (what we've been calling) the \noun{nlpaueb}
and \noun{customLegalBERT} models considered here, but two other legal
models as well. Their central finding (shared with most other LLM
model evaluations) is that bigger models with more parameters do better.
However, the models' ``legal prior'' (exposure to legal corpora)
also makes a difference. They also identify two interesting resources
for probing legal LLM, a set of criminal charges curated by FindLaw.com\footnote{https://www.findlaw.com/criminal/criminal-charges.html},
and the legal topics in the Wex dictionary\footnote{https://www.law.cornell.edu/wex/}
curated by the Legal Information Institute at Cornell University.
These can be used in a manner similar to the Dunn phrases in Section
\ref{subsec:Connecting-LLM-to-legal-language}, to connect LLM features
to known legal language.

A broadly available tokenizer specifically designed for special features
of legal texts (e.g., BlueBook citation styles) would do most to accelerate
work in legal AI. The tokenizers evaluated here, both the WordPiece
version used by \noun{BERT} and \noun{legalBert} and the SentencePiece
version used by \noun{customLegalBert} and \noun{nlpaueb}, are all
linguistically naive and seem to leave much useful language behind.
Very simple information retrieval techniques using corpus-wide token
frequencies are shown here to be helpful, for example separating common
``stop words'' from other tokens that are clear signifiers of legal
topics. 

There are other ways to embed natural language for LLM. Static, pre-computed
embedding techniques like Word2Vec\footnote{https://github.com/tensorflow/tensorflow/blob/r1.1/tensorflow/examples/tutorials/word2vec/}
and GloVe\footnote{https://nlp.stanford.edu/projects/glove/} are
often used. \citet{vallebueno2024statisticaluncertaintywordembeddings}
discuss an extension GloVe-V that allows characterization of the ``uncertainty''
associated with a word's embedding: ``We should be less certain about
a word\textquoteright s position in vector space the less data we
have on its co-occurrences in the raw text.'' \citet{nandi2024sneakingsyntaxtransformerlanguage}
describes an effort to capture some sentence syntax directly by imposing
an inductive bias on potential ``contextual spans'' of tokens.

Careful linguistic analysis, like that done by \citet{dunn2003judges}
on overruling judicial decision texts could be applied to other classes
of legal documents (e.g., judicial opinions discussing holdings of
other courts) to provide similarly useful insights.

\subsection{Limitations}

\begin{figure}
\begin{centering}
\includegraphics[height=2in]{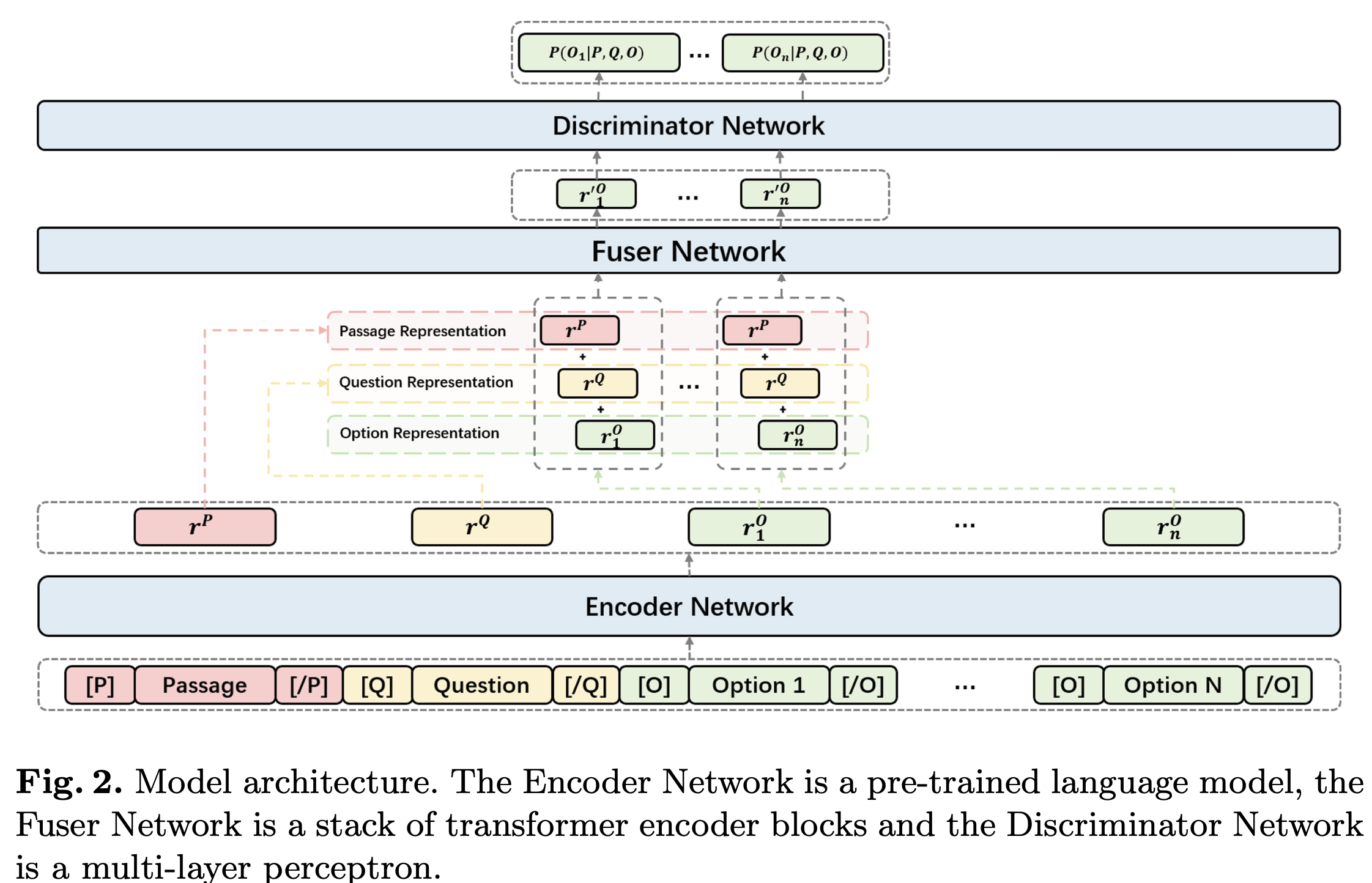}
\par\end{centering}
\caption{\label{fig:altMultChoiceModel}Alternative model for multiple choice
tasks \citet{Wang24} Figure 2}
\end{figure}

As discussed in Section \ref{subsec:CaseholdDS}, all models considered
here use the same SWAG \citep{zellers2018swag} construction for encoding
multiple choice questions that was applied to the \noun{casehold}
dataset. This construction only considers the text of the passage
against a single possible holding text, rather than \emph{across}
all possible options. A better reflection of the cognitive task facing
human decision makers is suggested by Figure \ref{fig:altMultChoiceModel}.
This multiple choice construction by \citet{Wang24} builds embeddings
over \emph{all} the textual components together, and then use a ``fuser''
network to capture interactions among options.

Publicly-available datasets and models are critical to support further
research in legal AI. As \citet{zhengguha2021} said four years ago: 
\begin{quote}
The Overruling dataset, for instance, required paying attorneys to
label each individual sentence. Once a company has collected that
information, it may not want to distribute it freely for the research
community. In the U.S. system much of this meta-data is hence retained
behind proprietary walls (e.g., Lexis and Westlaw), and the lack of
large-scale U.S. legal NLP datasets has likely impeded scientific
progress. 
\end{quote}
Experiments reported here have been constrained by modest computational
resources that allowed only fine-tuning of existing base models. 

Only integrated gradient attribution has been considered here. Providing
a baseline value for each example is crtitical, and only the default
(zero) values were used here. Establishing more appropriate baselines
is an important remaining task.

There are other attribution techniques, for example perturbation techniques
selectively ablating neural units, and instance-based techniques that
focus on specific example training instances. \citet{Yu2024RevealingTP}
considers instance-based vs \textquotedbl neuron-base\textquotedbl{}
methods like integrated gradient and also hybrids of them. Recent
theoretic work \citep{Bilodeau24} suggests IG and all other feature
attribution methods are ``impossible'': any feature attribution
method (including IG) that is complete and linear cannot do better
than random guessing. 

As early AI tools make their way into the legal market, it is critical
to doubt their reliability in real-life application \citep{magesh2024hallucinationfreeassessingreliabilityleading}.

\subsection{Conclusions}

All LLM require some method of embedding of language before learning
can proceed, and GIGO\footnote{Garbage In, Garbage Out} remains a
useful heuristic for this important new class of software. Our central
result is that attribution methods like integrated gradient can provide
useful insights into the behavior of legal LLM classifiers, related
specifically to features of the legal language and its tokenization.
It remains unclear why models with \emph{least} exposure to legal
texts best separate attribution scores of correct and incorrect examples.
As discussed in Section \ref{subsec:Variation-across-LLM}, it is
conceivable that ``mixture of expert'' techniques could combine
across models to attain accuracies near 80\%.

\texttt{\bibliographystyle{ACM-Reference-Format}
\bibliography{icail25}
 } 
\end{document}